\documentclass{article}
\usepackage{stywhispers,amsmath,epsfig}

\usepackage{url}
\usepackage{algorithm}
\usepackage{algorithmic}
\usepackage{multirow}
\newcommand{\tabincell}[2]{\begin{tabular}{@{}#1@{}}#2\end{tabular}}

\title{A Novel Statistical Metric Learning for Hyperspectral Image Classification}
\name{Zhiqiang Gong, Ping Zhong,  Weidong Hu, Zixuan Xiao, Xuping Yin\thanks{This work was supported in part by the NSF of China under Grant 61271439 and 61671456, the Program for New Century Excellent Talents in University under Grant NECT-13-0164, FANEDD under Grant 201243. (Corresponding Author: Ping Zhong. E-mail: zhongping@nudt.edu.cn)}}
\address{National Key Laboratory of Science and Technology on ATR, \\ College of Electrical Science and Techonology, National University of Defense Technology, \\ Changsha 410073, Hunan, China}
\begin{document}
%
\maketitle
\begin{abstract}
In this paper, a novel statistical metric learning is developed for spectral-spatial classification of the hyperspectral image. First, the standard variance of the samples of each class in each batch is used to decrease the intra-class variance within each class. Then, the distances between the means of different classes are used to penalize the inter-class variance of the training samples. Finally, the standard variance between the means of different classes is added as an additional  diversity term to repulse different classes from each other. Experiments have conducted over two real-world hyperspectral image datasets and the experimental results have shown the effectiveness of the proposed statistical metric learning.
\end{abstract}
\section{Introduction}
\label{sec:intro}

Recently, hyperspectral image classification has become a hot topic due to the plentiful information from the hundreds of spectral channels contained in the image \cite{zhong2018}. However, the great similarity occurred in the spectral bands of different objects makes the task be a challenge one. Moreover, the limited number of labelled samples in real-world applications increases the difficulty to obtain discriminative spectral features from the image. To overcome this problem, spatial information is usually incorporated into the representation to provide discriminative features. However, modelling the spatial and spectral directly with usual handcrafted features cannot capture the complex structure and high-level information within the image.

Deep models have shown powerful ability in describing the abstract and high-level information and presented remarkable performance in many computer vision tasks, such as the object detection, face recognition, as well as in the literature of hyperspectral image classification \cite{chen2015,chen2016}. Many deep models, such as the deep belief networks \cite{chen2015} and the convolutional neural networks \cite{chen2016,gong2019}, have been applied for the hyperspectral image processing tasks. However, due to the limited  and unbalanced  training samples in the hyperspectral image, general training process of deep model for the hyperspectral image usually makes the learned models be sub-optimal.

To overcome this problem, metric learning which tries to maximize the inter-class variance while minimizing the intra-class variance is usually applied in the training process of the deep model for the hyperspectral image \cite{gong2019}. Generally, the metric learning constructs the image pairs or triplet data to penalize the inter-class distance and the intra-class distance. By increasing the variance between samples from the same class and decreasing the variance between samples from different classes, the learned features can be more discriminative to separate different objects. However, general methods to implement the metric learning should construct the image pairs. Besides, the training process would be unbalanced due to the unbalance of training samples.

This work develops a novel statistical metric learning (SML) which increases the inter-class variance and decreases the intra-class variance from the statistical view. All the samples from the same class are looked as a distribution. The variance from each class is used to formulate the intra-class variance. The Euclidean distances between the sample means from different classes are used to measure the inter-class variance. Moreover, the variance between different sample means is added as a diversity regularization to repulse different classes from each other. The SML is easy to implement. Moreover, under the SML, the variance is measured from the class view which can balance the training process with unbalanced training samples.

Just as \cite{wen2016}, this work jointly learns the developed SML and the softmax loss for hyperspectral image classification. The softmax loss tries to take advantage of the point-to-point information while the SML makes use of the class-wise information and further improves the representational ability of the learned features. Experimental results over two commonly used hyperspectral image have demonstrated the effectiveness of the developed method.

\begin{figure*}[htp]

\begin{minipage}[b]{.98\linewidth}
  \centering
 \centerline{\epsfig{figure=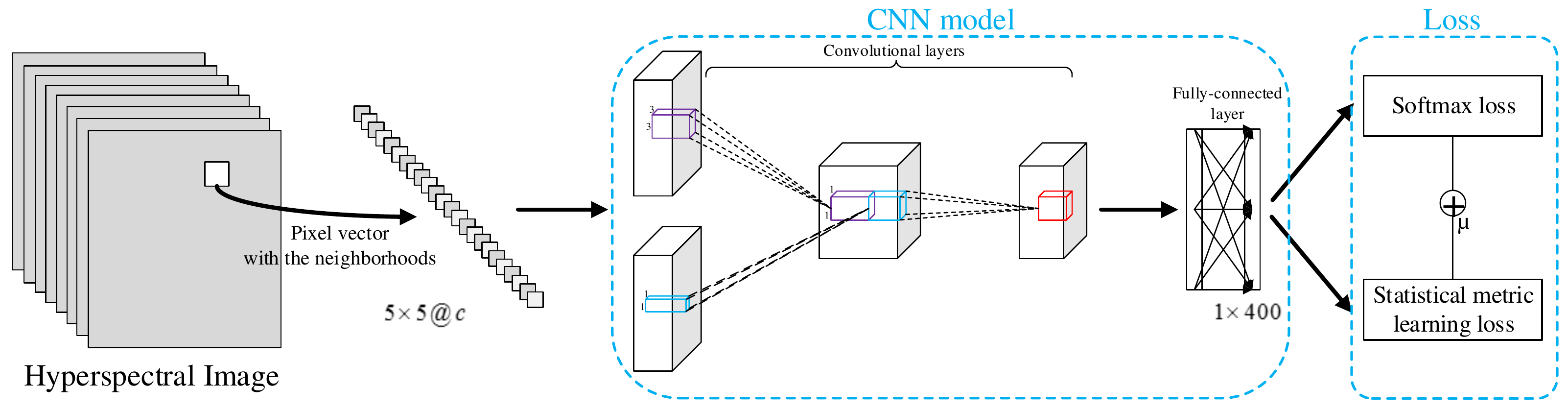,width=16cm}}
\end{minipage}
\hfill
\vspace{-0.5cm}
\caption{Flowchart of the proposed method for hyperspectral image classification. $a\times b@c$ describes the dimension of the data where a, b, and c represent the width, height and the channel of the data. $\mu$ denotes the loss weight of the proposed SML in the training process.}
\label{fig:flowchart}
\vspace{-0.5cm}
\end{figure*}

\section{Proposed Method}
\label{sec:format}

Let us denote $X=\{{\bf x}_1,{\bf x}_2,\cdots,{\bf x}_N\}$ as the set of training samples of the hyperspectral image where $N$ is the number of training samples and $y_i$ as the label of the sample ${\bf x}_i$.
$y_i \in Y=\{y_{m_1},y_{m_2},\cdots,y_{m_\Lambda}\}$ where $\Lambda$ is the number of the sample classes.

\subsection{General metric learning}
Since convolutional neural networks (CNNs) have presented impressive results in hyperspectral image classification \cite{li2017},
as Fig. \ref{fig:flowchart} shows, this work will choose the CNN model as \cite{gong2019} to extract features from the hyperspectral image.
To further improve the representational ability of the hyperspectral image, the metric learning is incorporated in the deep learning process. Generally, metric learning calculates the loss to measure the inter-class difference and intra-class similarity to decrease the intra-class variance and penalize the inter-class variance, simultaneously. Therefore, the loss can be formulated as
\begin{equation}\label{eq:01}
  L_0=L_{inter}+L_{intra},
\end{equation}
where $L_{inter}$ measures the penalization between different classes and $L_{intra}$ calculates the penalization within each class. Contrastive loss and triplet loss are the commonly used metric learning methods.

Contrastive loss constructs the image pairs $({\bf x}_i,{\bf x}_j)$ (including the positive pairs and the negative pairs) where the positive pair denotes images from the same class and the negative pair denotes images from different classes. The contrastive loss decreases the distances of the positive pairs and penalizes the negative pairs. It can be formulated as
\begin{equation}\label{eq:03}
  L_0=\sum_{({\bf x}_i, {\bf x}_j)}\{I(y_i=y_j)D({\bf x}_i, {\bf x}_j)+I(y_i\neq y_j)[\kappa-D({\bf x}_i, {\bf x}_j)]\}
\end{equation}
where $D({\bf x}_i, {\bf x}_j)=\|f({\bf x}_i)-f({\bf x}_j)\|$ and $f({\bf x}_i)$ is the feature of sample ${\bf x}_i$. $\kappa$ is the margin. $I(\cdot)$ represents the indicative function.

Triplet loss constructs the triplet data $({\bf x}_a,{\bf x}_p,{\bf x}_n)$ where $({\bf x}_a,{\bf x}_p)$ comes from the same class and $({\bf x}_a,{\bf x}_n)$ is from different classes. The loss is formulated based on the triplet data,
\begin{equation}\label{eq:03}
  L_0=\sum_{({\bf x}_a, {\bf x}_p, {\bf x}_n)}\{D({\bf x}_a, {\bf x}_p)+\kappa-D({\bf x}_a, {\bf x}_n)\}
\end{equation}

These former metric learning methods usually require data preprocessing to construct the image pairs or triplet data. Besides, these methods commonly consider the sample correlation and ignore the class correlation. This would negatively affected the classification performance over unbalanced data especially for hyperspectral image. Therefore, to overcome these problems, this work will develop a novel statistical metric learning.

\subsection{Statistical metric learning}

Given a training batch $B$. This work looks each class as a distribution and we will implement the metric learning from the statistical view.
Denote ${\bf z}_i$ as the feature of ${\bf x}_i$ extracted from the deep model.
$X^B_{k}=\{{\bf x}_{s_1},{\bf x}_{s_2},\cdots,{\bf x}_{s_{n_k}}\}$ represents the samples of the $k$-th class in the batch. Then, $C^B_k=\{{\bf z}_{s_1},{\bf z}_{s_2},\cdots,{\bf z}_{s_{n_k}}\}$ denotes the extracted features of the $k$-th class where $n_k$ is the number of the samples in the class.

The sample mean $\bar{C}_k$ of the $k$-th class in $B$ is calculated as
\begin{equation}\label{eq:04}
  \bar{C}_k=\frac{1}{n_k}\sum_{i=1}^{n_k}{\bf z}_{s_{i}}
\end{equation}
Then, the variance $I_k$ of the samples of  the $k$-th class in $B$ can be calculated as
\begin{equation}\label{eq:02}
  I_k=\frac{1}{n_k}\sum_{j=1}^{n_k}(\bar{{C}}_k-{\bf z}_j)^2
\end{equation}

Since the variance is a measure of how spread out  a data set is, this work will take advantage of
the variance of different classes in the batch $B$ to formulate the intra-class variance of the training batch. Then, $L_{intra}$ can be formulated as
\begin{equation}\label{eq:03}
  L_{intra}=\frac{1}{\Lambda}\sum_{k=1}^{\Lambda}I_k=\frac{1}{\Lambda}\sum_{k=1}^{\Lambda}\frac{1}{n_k}\sum_{j=1}^{n_k}(\bar{{C}}_k-{\bf z}_j)^2
\end{equation}
Besides, this work tries to enlarge the Euclidean distance between the sample means of different classes to enlarge the inter-class variance.
\begin{equation}\label{eq:03}
  L_{inter}=\frac{2}{\Lambda^2}\sum_{i\neq j}^{\Lambda}(\bar{C}_i-\bar{C}_j)^2
\end{equation}

Moreover,
denote $C_0$ as the center of all the classes in the batch
\begin{equation}\label{eq:03}
  C_0=\frac{1}{\Lambda}\sum_{k=1}^{\Lambda}\bar{C}_k
\end{equation}
The variance of all the means of different classes is calculated as a diversity regularization to repulse different classes to each other. It can be formulated as
\begin{equation}\label{eq:03}
  L_{diversity}=\frac{1}{\Lambda}\sum_{k=1}^{\Lambda}(C_0-\bar{C}_k)^2
\end{equation}

The statistical metric learning (SML) penalizes the variance of each class and the Euclidean distance between the sample means of different classes. Besides, the SML penalizes the variance between the sample means of different classes as the diversity term to repulse different classes from each other. Then, the loss can be formulated as
\begin{equation}\label{eq:03}
  L=\alpha L_{intra} - \beta L_{inter} - \lambda L_{diversity}
\end{equation}

The SML calculates the inter-class and intra-class variance between the samples in the training batch and is easy to implement. Moreover, the SML measures the difference from the class view which can solve the unbalanced training from unbalanced data.
It should also be noted that, as Fig. \ref{fig:flowchart} shows, this work jointly learns the softmax loss and the proposed SML to take advantage of both the point-to-point correlation and the class-wise correlation.

\subsection{Implementation of the proposed method}

The model is trained by the stochastic gradient descent method and back propagation is used for the training process of the proposed method \cite{back_propagation}. Generally, the main problem for the training process is to compute the learning loss w.r.t. the training samples.

The partial of the softmax loss w.r.t. ${\bf x}_i$ can be computed as Caffe which is the deep framework used in this work (see \cite{caffe} for details). The partial of $L_{intra}$ w.r.t. ${\bf x}_i$ can be computed by
\begin{equation}\label{eq:03}
  \frac{\partial L_{intra}}{\partial {\bf x}_i}=\frac{2}{\Lambda}\sum_{k=1}^{\Lambda}\frac{1}{n_k}I({\bf x}_i\in X^B_{k})({\bf x}_i-\bar{C}_k)
\end{equation}
Besides, the partial of $L_{inter}$ w.r.t. ${\bf x}_i$ can be calculated as
\begin{equation}\label{eq:03}
  \frac{\partial L_{inter}}{\partial {\bf x}_i}=\frac{4}{\Lambda^2} \sum_{k=1}^{\Lambda}\frac{1}{n_k}I({\bf x}_i\in X^B_{k})\sum_{l\neq k}^{\Lambda}(\bar{C}_k-\bar{C}_l)
\end{equation}
The partial of $L_{diversity}$ w.r.t. ${\bf x}_i$ can be calculated as
\begin{equation}\label{eq:03}
  \frac{\partial L_{diversity}}{\partial {\bf x}_i}=\frac{2}{\Lambda}\sum_{k=1}^{\Lambda}\frac{1}{n_k}I({\bf x}_i\in X^B_{k})(\bar{C}_k-C_0)
\end{equation}
Through back propagation with the former equations, the CNN model can be trained and discriminative features can be learned from the hyperspectral image.

\section{Experiments}\label{sec:pagestyle}

\begin{figure}[t]

\begin{minipage}[b]{.36\linewidth}
  \centering
 \centerline{\epsfig{figure=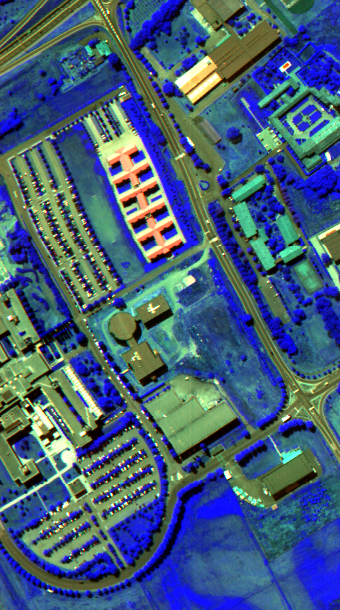,width=3cm}}
  \centerline{(a)}\medskip
\end{minipage}
\hfill
\begin{minipage}[b]{0.36\linewidth}
  \centering
 \centerline{\epsfig{figure=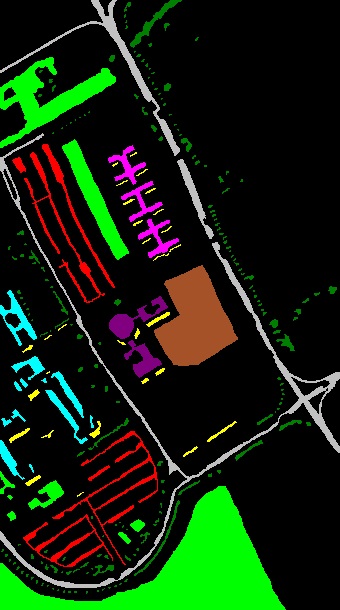,width=3cm}}
  \centerline{(b)}\medskip
\end{minipage}
\hfill
\begin{minipage}[b]{.24\linewidth}
  \centering
 \centerline{\epsfig{figure=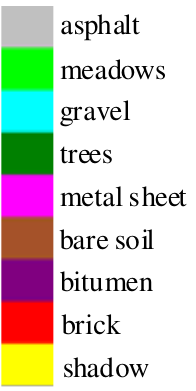,width=2cm}}
  \centerline{(c)}\medskip
\end{minipage}
\hfill

\vspace{-0.5cm}
\caption{Pavia university dataset. (a) False color composite (band 10, 60, 90); (b) ground truth; (c) map color.}
\label{fig:pavia}
\vspace{-0.3cm}
\end{figure}

\begin{figure}[t]

\begin{minipage}[b]{.36\linewidth}
  \centering
 \centerline{\epsfig{figure=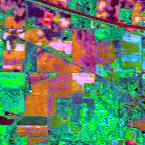,width=3cm}}
  \centerline{(a)}\medskip
\end{minipage}
\hfill
\begin{minipage}[b]{0.36\linewidth}
  \centering
 \centerline{\epsfig{figure=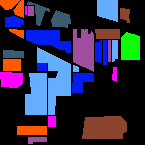,width=3cm}}
  \centerline{(b)}\medskip
\end{minipage}
\hfill
\begin{minipage}[b]{.24\linewidth}
  \centering
 \centerline{\epsfig{figure=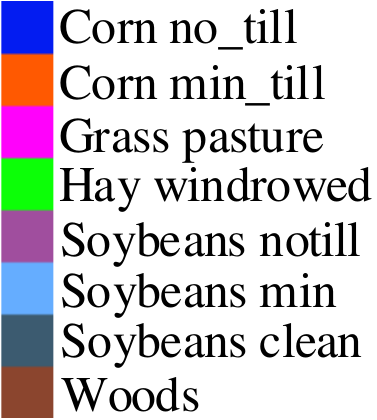,width=2cm}}
  \centerline{(c)}\medskip
\end{minipage}
\hfill

\vspace{-0.5cm}
\caption{Indian Pines dataset. (a) False color composite (band 30, 60, 90); (b) ground truth; (c) map color.}
\label{fig:indian}
\vspace{-0.6cm}
\end{figure}

To further validate the effectiveness of the proposed statistical metric learning, this work conducts experiments over commonly used hyperspectral images, namely the Pavia University and the Indian Pines, and further compares the developed metric learning with other methods.
Pavia University \cite{pavia} consists of $610 \times 340$ pixels with 115 bands ranging from 0.43 to 0.86 ${\mu}m$ (See Fig. \ref{fig:pavia} for details). 103 channels are used for experiments due to the noise. 42,776 labelled samples which are divided into nine classes are selected.
Indian Pines \cite{indian} consists of $145 \times 145$ pixels with 224 spectral channels raning from 0.4 to 2.45 ${\mu}m$ (See Fig. \ref{fig:indian} for details). 24 spectral bands are removed due to the noise and the remainder are used for experiments. A total of 8598 labelled samples from eight classes are chosen from the image. In the experiments, 200 samples of each class are used for training and the remainder for testing.

Caffe \cite{caffe} is chosen to implement the deep learning framework. In the experiments, the learning rate and the training epoch is set to 0.001 and 40000, respectively. The tradeoff parameters $\alpha, \beta,$ and $\lambda$ are set to 1, 0.01 and 0.001, separately. The loss weight of SML in the training process is set to 0.0002.
The experimental results in the paper are from the mean and standard variance of ten runs of training and testing.


\subsection{Classification Results}

The classification results of the proposed method over Pavia University and Indian Pines are shown in table \ref{table:pavia} and \ref{table:indian}, respectively. From table \ref{table:pavia}, we can find that the CNN model which trained with general softmax loss can obtain $98.72\pm 0.27$ over Pavia University dataset while the performance can achieve $99.39\pm 0.09$ when trained with the proposed SML. As table \ref{table:indian} shows, for Indian Pines dataset, the CNN model can provide an accuracy of $98.46\pm 0.20$ and $98.94\pm 0.30$, respectively. In conclusion, the proposed method can significantly improve the representational ability of the CNN model for hyperspectral image classification.

\begin{table}[t]
\begin{center}
\caption{Classification accuracies (\%) of CNN and the proposed method over Pavia University. The results from CNN are trained with only Softmax loss.}
\label{table:pavia}
\vspace{1ex}
\begin{tabular}{c | c | c c}
\hline
\multicolumn{2}{c|}{\bf Methods}     &  {\bf CNN} &  {\bf Proposed Method}  \\
\hline\hline
\multirow{9}{*}{\rotatebox{90}{\tabincell{c}{\textbf{Classification} \\ \textbf{Accuracies (\%)}}}}          & C1   &  $98.59\pm 0.26$  &  $99.31\pm 0.37$ \\
                                                                                                             & C2   &  $98.84\pm 0.62$  &  $99.80\pm 0.11$ \\
                                                                                                             & C3   &  $95.57\pm 1.43$  &  $96.58\pm 1.05$ \\
                                                                                                             & C4   &  $98.76\pm 1.28$  &  $99.08\pm 0.64$ \\
                                                                                                             & C5   &  $100.0\pm 0.00$  &  $100.0\pm 0.00$ \\
                                                                                                             & C6   &  $99.81\pm 0.25$  &  $99.59\pm 0.53$ \\
                                                                                                             & C7   &  $99.15\pm 0.21$  &  $99.82\pm 0.27$ \\
                                                                                                             & C8   &  $97.62\pm 1.20$  &  $98.48\pm 0.70$ \\
                                                                                                             & C9   &  $100.0\pm 0.00$  &  $100.0\pm 0.00$ \\
 \hline
 \multicolumn{2}{c|}{{\bf OA}  (\%)}      &  $98.72\pm 0.27$ &  $99.39\pm 0.09$  \\
 \hline
 \multicolumn{2}{c|}{{\bf AA}  (\%)}      &  $98.70\pm 0.17$ &  $99.19\pm 0.16$  \\
 \hline
 \multicolumn{2}{c|}{{\bf KAPPA} (\%)}    &  $98.27\pm 0.36$ &  $99.19\pm 0.12$  \\
\hline
\end{tabular}
\end{center}
\vspace{-1cm}
\end{table}

\begin{table}[htp]
\vspace{-0.6cm}
\begin{center}
\caption{Classification accuracies (\%) of CNN and the proposed method over Indian Pines.}
\label{table:indian}
\vspace{1ex}
\begin{tabular}{c | c | c c}
\hline
\multicolumn{2}{c|}{\bf Methods}       &  {\bf CNN} &  Proposed Method  \\
\hline\hline
\multirow{9}{*}{\rotatebox{90}{\tabincell{c}{\textbf{Classification} \\ \textbf{Accuracies (\%)}}}}          & C1   &  $98.12\pm 0.87$  &  $98.20\pm 0.59$ \\
                                                                                                             & C2   &  $99.84\pm 0.19$  &  $99.81\pm 0.30$ \\
                                                                                                             & C3   &  $100.0\pm 0.00$  &  $99.93\pm 0.14$ \\
                                                                                                             & C4   &  $100.0\pm 0.00$  &  $100.0\pm 0.00$ \\
                                                                                                             & C5   &  $99.40\pm 0.57$  &  $99.23\pm 0.51$ \\
                                                                                                             & C6   &  $96.57\pm 0.96$  &  $98.11\pm 0.79$ \\
                                                                                                             & C7   &  $99.71\pm 0.49$  &  $99.90\pm 0.17$ \\
                                                                                                             & C8   &  $99.98\pm 0.04$  &  $99.88\pm 0.22$ \\
 \hline
 \multicolumn{2}{c|}{{\bf OA}  (\%)}      &  $98.46\pm 0.20$ &  $98.94\pm 0.30$  \\
 \hline
 \multicolumn{2}{c|}{{\bf AA}  (\%)}      &  $99.22\pm 0.02$ &  $99.38\pm 0.15$  \\
 \hline
 \multicolumn{2}{c|}{{\bf KAPPA} (\%)}    &  $98.10\pm 0.24$ &  $98.70\pm 0.36$  \\
\hline
\end{tabular}
\end{center}
\vspace{-0.8cm}
\end{table}

\begin{figure}[htp]

\begin{minipage}[b]{0.32\linewidth}
  \centering
 \centerline{\epsfig{figure=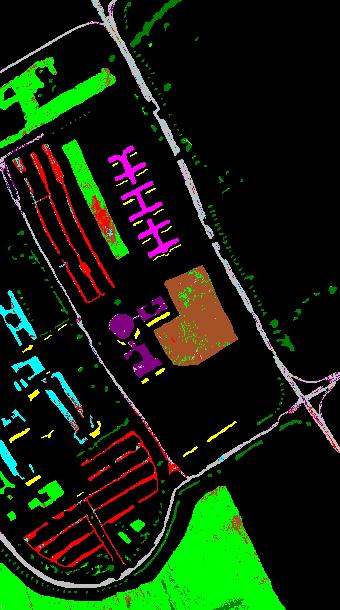,width=2.6cm}}
  \centerline{(a)}\medskip
\end{minipage}
\hfill
\begin{minipage}[b]{.32\linewidth}
  \centering
 \centerline{\epsfig{figure=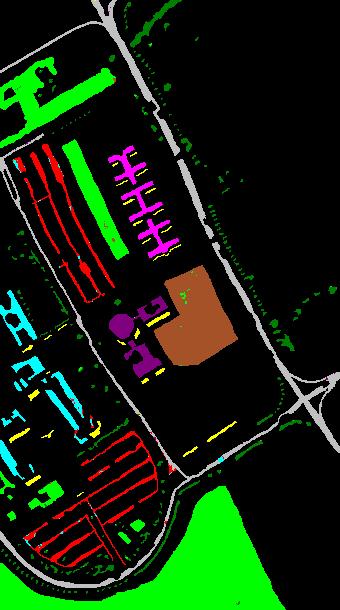,width=2.6cm}}
  \centerline{(b)}\medskip
\end{minipage}
\hfill
\begin{minipage}[b]{0.32\linewidth}
  \centering
 \centerline{\epsfig{figure=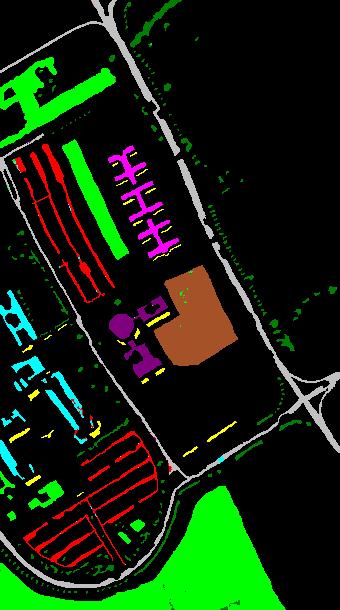,width=2.6cm}}
  \centerline{(c)}\medskip
\end{minipage}
\vspace{-0.5cm}
\caption{Classification maps of Pavia University obtained by (a) SVM;  (b) CNN; (c) Proposed Method.}
\label{fig:pavia_maps}
\vspace{-0.6cm}
\end{figure}

\begin{figure}[htp]

\begin{minipage}[b]{0.32\linewidth}
  \centering
 \centerline{\epsfig{figure=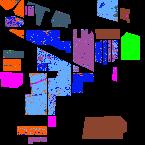,width=2.6cm}}
  \centerline{(a)}\medskip
\end{minipage}
\hfill
\begin{minipage}[b]{.32\linewidth}
  \centering
 \centerline{\epsfig{figure=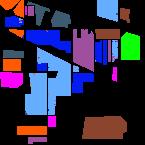,width=2.6cm}}
  \centerline{(b)}\medskip
\end{minipage}
\hfill
\begin{minipage}[b]{0.32\linewidth}
  \centering
 \centerline{\epsfig{figure=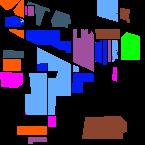,width=2.6cm}}
  \centerline{(c)}\medskip
\end{minipage}
\vspace{-0.5cm}
\caption{Classification maps of Indian Pines obtained by (a) SVM;  (b) CNN; (c) Proposed Method.}
\label{fig:indian_maps}
\vspace{-0.8cm}
\end{figure}

Besides, Fig. \ref{fig:pavia_maps} and \ref{fig:indian_maps} also show the classification maps of SVM, CNN, and the proposed method over Pavia University and Indian Pines, separately. By comparisons of Figs. \ref{fig:pavia_maps} and \ref{fig:indian_maps}, it can be noted that, the deep model can significantly improve the performance. Besides, the proposed method can further improve the representational ability of the CNN model and discriminate the objects with highly overlappings.

\subsection{Classification Performance With Different Number of Training Samples}

\begin{figure}[htp]

\begin{minipage}[b]{.48\linewidth}
  \centering
 \centerline{\epsfig{figure=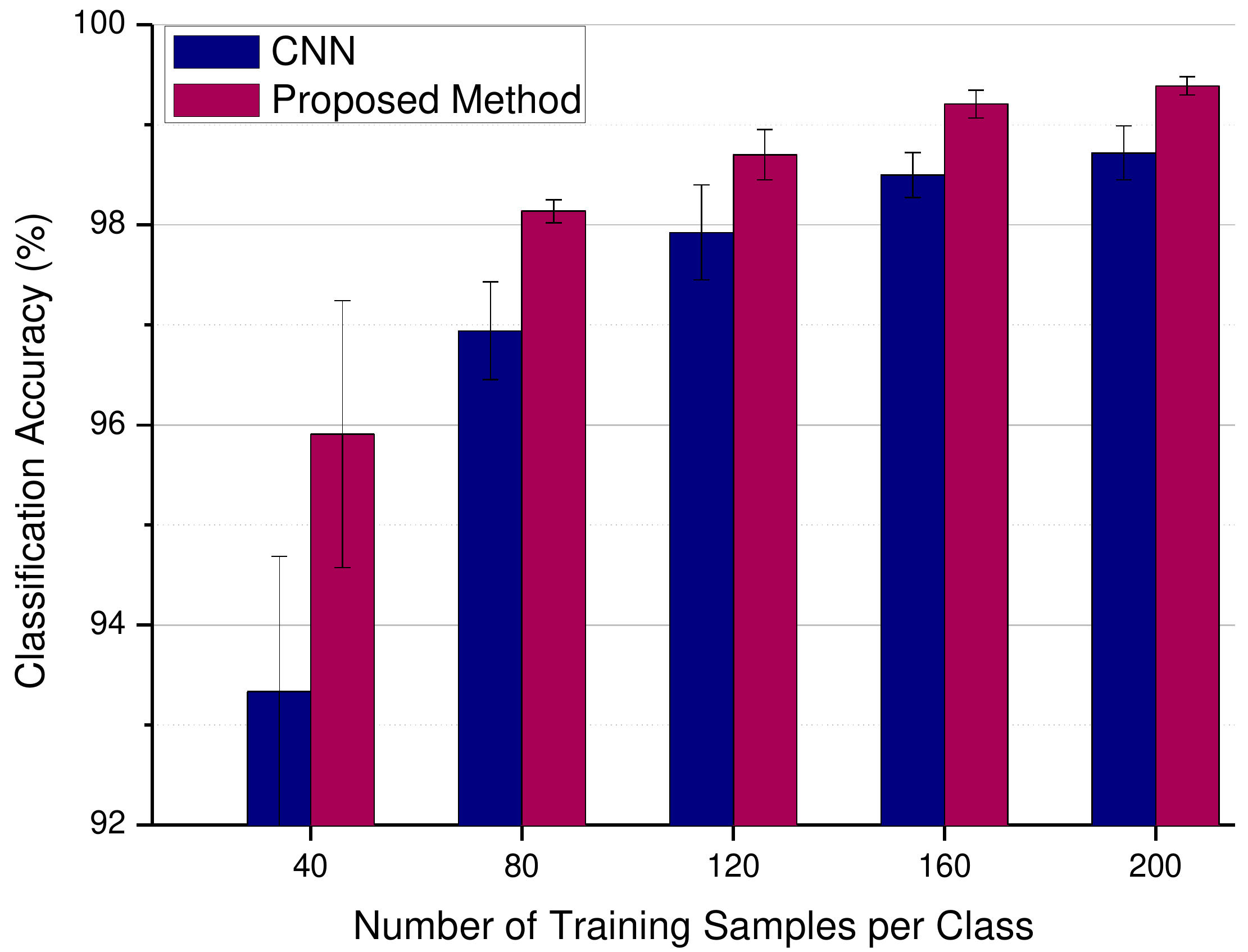,width=4.0cm}}
  \centerline{(a) Ucmerced Land Use}\medskip
\end{minipage}
\hfill
\begin{minipage}[b]{0.48\linewidth}
  \centering
 \centerline{\epsfig{figure=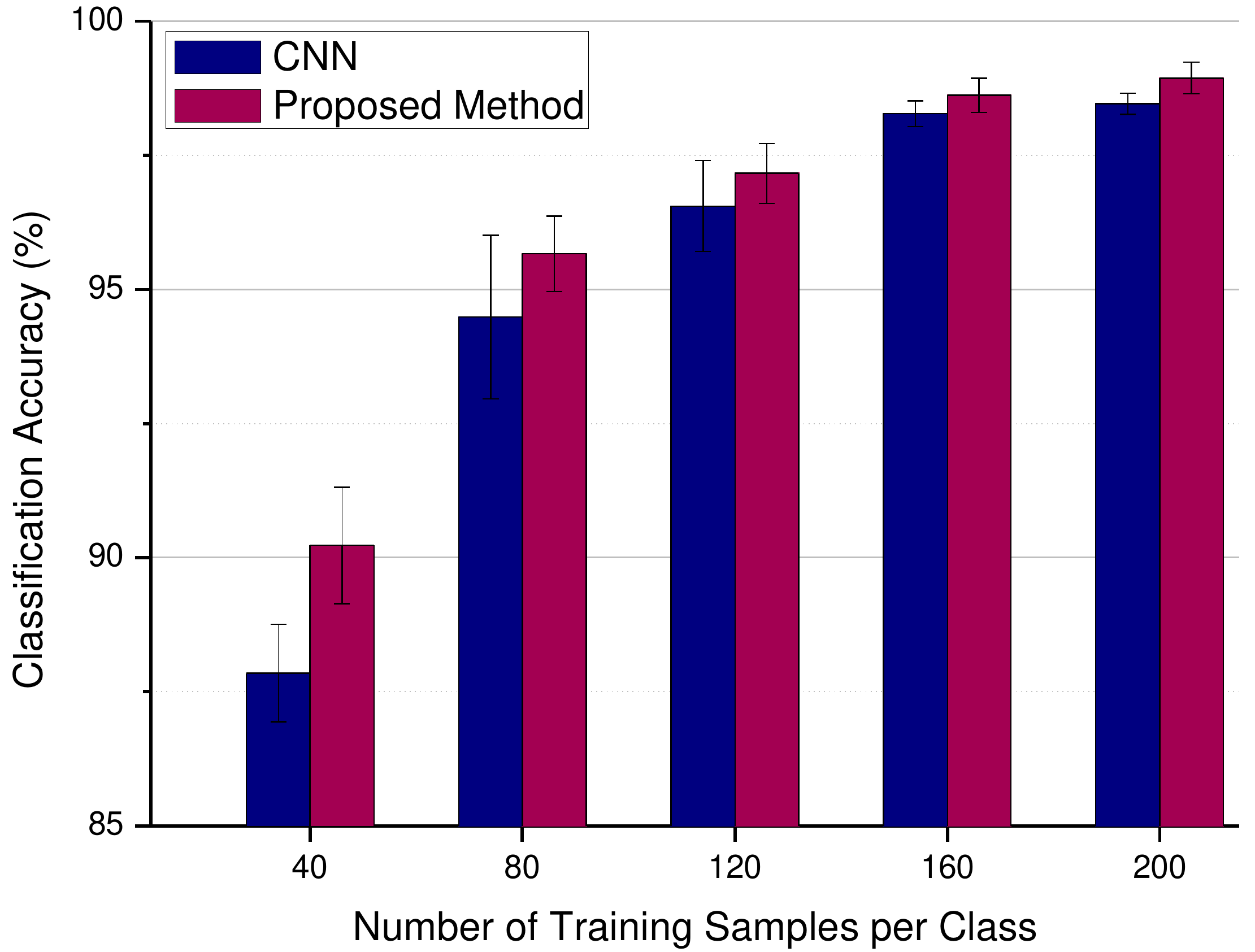,width=4.0cm}}
  \centerline{(b) Indian Pines}\medskip
\end{minipage}
\vspace{-0.6cm}
\caption{Classification Results of the Proposed method with different number of training samples per classification over different datasets.}
\label{fig:number}
\vspace{-0.6cm}
\end{figure}

To further validate the performance of the proposed method, in Fig. \ref{fig:number} we provide the classification accuracies of the CNN model with general softmax loss and the proposed SML over Pavia University and Indian Pines, respectively. It can be noted that the proposed SML method obtains better performance with different number of training samples. Interestingly, the proposed SML can provide a large improvement when compared with model trained with general softmax loss with less training samples. Since the SML considers the correlation from the class view, the number of samples shows less effects on the performance of SML. However, less samples can significantly affect the performance of softmax loss. Therefore, the SML can play a more important role in the performance with less training samples.

\subsection{Comparisons with other State-of-the-Art Methods}

To comprehensively show the effectiveness of the proposed method, we compare the developed method with several state-of-the-art methods.

Compared with shallow methods, we can find that the proposed method shows better performance than SIFT \cite{zhong2018}. Compared with deep methods, it can be noted from table \ref{table:comparison_pavia} that the proposed method obtains better performance over Pavia University when compared with D-DBN-PF ($93.11\pm 0.06$) \cite{zhong2018} and CNN ($98.72\pm 0.27$) \cite{gong2019} which are generally used deep model. Besides, from table \ref{table:comparison_indian}, we can also find that the proposed method achieve better performance over Indian Pines when compared with D-DBN-PF ($91.03\pm 0.12$)\cite{zhong2018} and CNN ($98.46\pm 0.20$)\cite{gong2019}. In conclusion, the proposed method can provide better performance when compared with both the shallow and deep methods over the hyperspectral image.

\begin{table}[t]
\begin{center}
\caption{Classification accuracy (Mean $\pm$ SD) obtained by different methods on Pavia University.}
\label{table:comparison_pavia}
\vspace{1ex}
\begin{tabular}{l c}
\hline
{\bf Method}     &    {\bf Accuracy (\%)}  \\
\hline\hline
SIFT\cite{zhong2018}    &  $90.73\pm 0.16$   \\
D-DBN-PF\cite{zhong2018} &  $93.11\pm 0.06$  \\
CNN\cite{gong2019}  &  $98.72\pm 0.27$  \\
{\bf Proposed Method}   &    $99.39\pm 0.09 $      \\
\hline
\end{tabular}
\end{center}
\vspace{-0.8cm}
\end{table}

\begin{table}[t]
\begin{center}
\vspace{-0.8cm}
\caption{Classification accuracy (Mean $\pm$ SD) obtained by different methods on Indian Pines.}
\label{table:comparison_indian}
\vspace{1ex}
\begin{tabular}{l c}
\hline
{\bf Method}     &    {\bf Accuracy (\%)}  \\
\hline\hline
SIFT\cite{zhong2018}    &  $87.65\pm 0.15$   \\
D-DBN-PF\cite{zhong2018} &  $91.03\pm 0.12$  \\
CNN\cite{gong2019}  &   $98.46\pm 0.20$ \\
{\bf Proposed Method}   &    $98.94\pm 0.30$      \\
\hline
\end{tabular}
\end{center}
\vspace{-0.8cm}
\end{table}

\section{Conclusions}
\label{sec:typestyle}

This work develops a novel statistical metric learning for hyperspectral image classification. The developed SML takes advantage of the sample variance of each class to calculate the intra-class variance. Moreover, the distances between the mean value of samples of each class are used to penalize the inter-class variance. In addition, the variance of the  means of different classes is added as additional  diversity regularization to repulse different classes from each other. Experimental results have demonstrated that the proposed method achieves better performance when compared with other state-of-the-art methods.

In further work, we would like to apply the proposed method to other remote sensing datasets. Moreover, other statistics which can measure the variance of the distribution is another direction to improve the performance of general deep learning.


\bibliographystyle{IEEEbib}
\bibliography{strings,refs}

\end{document}